\PassOptionsToPackage{unicode}{hyperref}
\PassOptionsToPackage{hyphens}{url}
\PassOptionsToPackage{dvipsnames,svgnames,x11names}{xcolor}
\documentclass[
  12pt]{article}

\usepackage{amsmath,amssymb}
\usepackage{multirow}
\usepackage{iftex}
\ifPDFTeX
  \usepackage[T1]{fontenc}
  \usepackage[utf8]{inputenc}
  \usepackage{textcomp} 
\else 
  \usepackage{unicode-math}
  \defaultfontfeatures{Scale=MatchLowercase}
  \defaultfontfeatures[\rmfamily]{Ligatures=TeX,Scale=1}
\fi
\usepackage{lmodern}
\ifPDFTeX\else  
\fi
\IfFileExists{upquote.sty}{\usepackage{upquote}}{}
\IfFileExists{microtype.sty}{
  \usepackage[]{microtype}
  \UseMicrotypeSet[protrusion]{basicmath} 
}{}
\makeatletter
\@ifundefined{KOMAClassName}{
  \IfFileExists{parskip.sty}{%
    \usepackage{parskip}
  }{
    \setlength{\parindent}{0pt}
    \setlength{\parskip}{6pt plus 2pt minus 1pt}}
}{
  \KOMAoptions{parskip=half}}
\makeatother
\usepackage{xcolor}
\makeatletter
\ifx\paragraph\undefined\else
  \let\oldparagraph\paragraph
  \renewcommand{\paragraph}{
    \@ifstar
      \xxxParagraphStar
      \xxxParagraphNoStar
  }
  \newcommand{\xxxParagraphStar}[1]{\oldparagraph*{#1}\mbox{}}
  \newcommand{\xxxParagraphNoStar}[1]{\oldparagraph{#1}\mbox{}}
\fi
\ifx\subparagraph\undefined\else
  \let\oldsubparagraph\subparagraph
  \renewcommand{\subparagraph}{
    \@ifstar
      \xxxSubParagraphStar
      \xxxSubParagraphNoStar
  }
  \newcommand{\xxxSubParagraphStar}[1]{\oldsubparagraph*{#1}\mbox{}}
  \newcommand{\xxxSubParagraphNoStar}[1]{\oldsubparagraph{#1}\mbox{}}
\fi
\makeatother

\usepackage{longtable,booktabs,array}
\usepackage{calc} 
\usepackage{etoolbox}
\makeatletter
\patchcmd\longtable{\par}{\if@noskipsec\mbox{}\fi\par}{}{}
\makeatother
\IfFileExists{footnotehyper.sty}{\usepackage{footnotehyper}}{\usepackage{footnote}}
\makesavenoteenv{longtable}
\usepackage{graphicx}
\makeatletter
\def\maxwidth{\ifdim\Gin@nat@width>\linewidth\linewidth\else\Gin@nat@width\fi}
\def\maxheight{\ifdim\Gin@nat@height>\textheight\textheight\else\Gin@nat@height\fi}
\makeatother
\setkeys{Gin}{width=\maxwidth,height=\maxheight,keepaspectratio}
\makeatletter
\def\fps@figure{htbp}
\makeatother

\makeatletter
\@ifpackageloaded{caption}{}{\usepackage{caption}}
\AtBeginDocument{%
\ifdefined\contentsname
  \renewcommand*\contentsname{Table of contents}
\else
  \newcommand\contentsname{Table of contents}
\fi
\ifdefined\listfigurename
  \renewcommand*\listfigurename{List of Figures}
\else
  \newcommand\listfigurename{List of Figures}
\fi
\ifdefined\listtablename
  \renewcommand*\listtablename{List of Tables}
\else
  \newcommand\listtablename{List of Tables}
\fi
\ifdefined\figurename
  \renewcommand*\figurename{Figure}
\else
  \newcommand\figurename{Figure}
\fi
\ifdefined\tablename
  \renewcommand*\tablename{Table}
\else
  \newcommand\tablename{Table}
\fi
}
\@ifpackageloaded{float}{}{\usepackage{float}}
\floatstyle{ruled}
\@ifundefined{c@chapter}{\newfloat{codelisting}{h}{lop}}{\newfloat{codelisting}{h}{lop}[chapter]}
\floatname{codelisting}{Listing}

\makeatother
\makeatletter
\@ifpackageloaded{caption}{}{\usepackage{caption}}
\@ifpackageloaded{subcaption}{}{\usepackage{subcaption}}
\makeatother

\ifLuaTeX
  \usepackage{selnolig}  
\fi
\usepackage[]{natbib}
\usepackage{bookmark}

\IfFileExists{xurl.sty}{\usepackage{xurl}}{} 
\hypersetup{
  pdftitle={Linking Deep Convolutional Features to Interpretable Image Statistics: A Case Study on Sex Estimation From Shoeprints},
  pdfauthor={Jinyi Niu; Weining Shen},
  pdfkeywords={shoeprint analysis; convolutional neural network; transfer learning; feature interpretability; forensic image analysis},
  colorlinks=true,
  linkcolor={blue},
  filecolor={Maroon},
  citecolor={Blue},
  urlcolor={Blue},
  pdfcreator={LaTeX via pandoc}}

\newcommand{\anon}{1}

\begin{document}

\def\spacingset#1{\renewcommand{\baselinestretch}%
{#1}\small\normalsize} \spacingset{1}


\if1\anon
{
  \title{\bf Sex Estimation from Footwear Outsole Impressions Using CNN Transfer Learning and Interpretable Image Statistics}
\author{
  {Jinyi Niu} \\
  School of Mathematical Sciences, Fudan University \\
  Shanghai 200433, China \\
  \texttt{24210180054@fudan.edu.cn} \\[1.5ex]
   Ziyi Song \\
  Department of Statistics, University of California, Irvine \\
  Irvine, CA 92697-1250, USA \\
  \texttt{ziyis9@uci.edu} \\[1ex]
  Weining Shen\textsuperscript{$\ast$} \\
  Department of Statistics, University of California, Irvine \\
  Irvine, CA 92697-1250, USA \\
  \texttt{weinings@uci.edu} \\[1ex]
  \textsuperscript{$\ast$}Corresponding author.
}
  \maketitle
} \fi

\if0\anon
{
  \bigskip
  \bigskip
  \bigskip
  \begin{center}
    {\Large \bf Sex Estimation from Footwear Outsole Impressions Using CNN Transfer Learning and Interpretable Image Statistics}
\end{center}
  \medskip
} \fi

\bigskip
 \begin{abstract}

Footwear outsole impressions are a common form of forensic pattern evidence, yet quantitative methods for estimating wearer attributes from these images remain relatively underdeveloped. We investigate binary sex estimation from footwear outsole impressions by comparing convolutional neural network (CNN) transfer learning with traditional feature-based classification. Using a publicly available outsole-impression dataset, we adopt a shoe-level training and test partition that keeps replicate scans of the same physical shoe together to reduce data leakage. We evaluate pretrained CNNs through end-to-end fine-tuning, frozen feature extraction followed by support vector machine classification, and hybrid feature fusion incorporating handcrafted, geometric, and metadata-derived descriptors. Fine-tuned CNNs achieve the strongest overall predictive performance and substantially outperform traditional classifiers trained on the manually specified descriptors alone, while frozen-feature approaches offer a less computationally demanding alternative. Exploratory analysis of low-dimensional CNN representations reveals associations with frequency threshold ratio, image contrast, and wavelet-based summaries, providing a connection between learned representations and measurable properties of outsole impressions. These findings suggest that CNN transfer learning captures discriminative information beyond the descriptors considered and offers a promising approach to footwear-based forensic screening. Further validation on independently collected and casework-like impressions is needed before operational use.

\end{abstract}

\noindent%
\vfill

\newpage
\spacingset{1.8} 

\section{Introduction}

Footwear impressions are a common form of pattern evidence in forensic investigations and can provide useful information about a crime scene, the class characteristics of a shoe, and potential links between footwear evidence and persons of interest \citep{wu2022crime}. However, drawing statistically defensible conclusions from forensic images remains challenging. \citet{kafadar2024challenges} emphasize that many forms of pattern evidence have historically relied on examiner judgment, and that progress toward quantitative image-based frameworks has been uneven across forensic disciplines. Compared with areas such as biometric recognition, ballistics, and bloodstain pattern analysis, footwear evidence remains relatively underdeveloped from a quantitative modeling perspective. Within this broader setting, sex estimation from footwear impressions represents a specific investigative task: the goal is not to identify the shoe or the wearer, but to estimate a recorded binary sex attribute that may help narrow candidate pools during early-stage investigation. Prior criminological research has examined associations between sex and offense patterns \citep{nagel1983gender}, motivating the potential value of such attribute-level information as an investigative aid.

Traditional forensic shoeprint examination relies on trained examiners who compare visible characteristics such as outsole size, shape, tread design, and wear marks. These methods remain central to practice, but they can be time-consuming and may depend on image quality, examiner experience, and subjective interpretation, which has motivated computational approaches that extract reproducible quantitative information from footwear impressions. Early computational approaches typically followed a feature-engineering pipeline: image acquisition, preprocessing, feature extraction, and classification. Handcrafted features have described geometric properties such as length, width, area, and aspect ratio; texture properties such as Gabor-filter or Local Binary Pattern summaries; and statistical properties from intensity histograms or frequency-domain representations, and have been used as inputs to classical classifiers including support vector machines \citep{cortes1995support}, k-nearest neighbors \citep{peterson2009k}, decision trees \citep{kotsiantis2013decision}, and artificial neural networks \citep{yegnanarayana2009artificial}. Such approaches are appealing because the extracted variables are often interpretable and linkable to visible impression characteristics, but their performance is limited by the quality and completeness of manually specified descriptors, which may fail to capture subtle spatial, textural, and structural variation in outsole images. In parallel, the forensic statistics literature has developed principled approaches for evaluating footwear evidence, including a Bayesian hierarchical framework \citep{spencer2020bayesian} and public datasets such as the two-dimensional outsole-impression database \citep{park2020database} and the ShoeCase mock crime-scene footwear dataset \citep{tibben2023shoecase}. These contributions provide an important foundation for quantitative reasoning about footwear evidence, but much of this work has relied on handcrafted descriptors or structured statistical summaries, with comparatively limited use of modern representation learning methods for extracting information directly from raw shoeprint images.


Convolutional neural networks (CNNs) offer a complementary approach. Rather than requiring analysts to specify all relevant image descriptors in advance, CNNs learn hierarchical image representations directly from pixel data. This ability to learn features automatically has led to strong performance in image classification, object detection, and segmentation, and has motivated the use of deep learning in forensic image analysis. In the footwear domain, \citet{hassan2021deep}  proposed a deep learning approach for age prediction from shoeprints and also considered sex classification, introducing the ShoeNet model and reporting 86.07\% accuracy for sex classification from shoeprint pressure distributions. Other studies have used CNNs for shoeprint retrieval and matching: \citet{ma2019shoe} proposed a multi-part weighted CNN for shoeprint image retrieval, and \citet{zhang2017adapting}  studied the adaptation of CNNs to forensic shoeprint retrieval under limited labeled data. \citet{rida2019forensic} reviewed forensic shoeprint identification methods, including deep-learning-based approaches, and \citet{hassan2024deep} studied deep learning for shoeprint reconstruction from partial or degraded impressions.

These studies suggest that CNNs can extract discriminative information from footwear images, but two issues remain especially important for forensic applications. First, model evaluation must avoid data leakage. This is particularly relevant for datasets containing repeated scans of the same physical shoe: if scans of the same shoe appear in both training and test sets, a model may learn shoe-specific visual patterns rather than characteristics associated with the target attribute, leading to inflated estimates of predictive performance. Second, CNNs are often difficult to interpret. A model may achieve high accuracy, but its learned features are not directly expressed in terms of measurable image properties. This “black-box” concern is especially important in forensic contexts, where computational conclusions should ideally be transparent, reproducible, and connected to observable properties of the evidence \citep{kafadar2024challenges}.

Our paper addresses these issues by combining predictive benchmarking with exploratory representation analysis. We analyze the public footwear outsole impression dataset of \citet{park2020database}, which contains 1,500 scans from 150 pairs of used shoes, with five replicate scans per individual shoe. To reduce leakage across replicate impressions, all model comparisons are conducted using a shoe-level train/test partition. We compare four ImageNet-pretrained CNN architectures, VGG16 \citep{simonyan2014very}, ResNet-50 \citep{he2016deep}, EfficientNet-B0 \citep{tan2019efficientnet}, and MobileNet-V2 \citep{sandler2018mobilenetv2}, under three transfer learning strategies: end-to-end fine tuning, frozen CNN feature extraction followed by support vector machine classification, and hybrid feature fusion combining CNN-derived representations with handcrafted, geometric, and metadata-derived descriptors. To provide traditional baselines, we also evaluate support vector machines \citep{cortes1995support}, random forests \citep{breiman2001random}, and XGBoost classifiers \citep{chen2016xgboost} trained only on the auxiliary feature set.

Our study makes three contributions. First, it provides a leakage-aware empirical benchmark for sex estimation from footwear outsole impressions, comparing CNN-based transfer learning pipelines with traditional feature-based classifiers on the same held-out shoes. Second, it evaluates how different transfer learning strategies trade off predictive performance and computational cost, including both full fine tuning and more efficient frozen-feature approaches. Third, it conducts an exploratory representation analysis to examine whether CNN-derived feature spaces are associated with quantifiable descriptors such as contrast, frequency threshold ratio, footprint length, wavelet-based summaries, shoe brand, and shoe size. This analysis is not intended to provide a causal explanation of CNN decisions, but it offers a first step toward relating high dimensional learned representations to measurable properties of outsole impressions.

The remainder of the paper is organized as follows. Section~\ref{sec: data_description} describes the footwear outsole impression dataset, the binary sex estimation target, image preprocessing, and the shoe-level train/test partition. Section~\ref{sec: methods} presents the CNN transfer learning strategies, handcrafted and metadata-derived features, traditional baselines, UMAP-based representation analysis, and evaluation metrics. Section~\ref{sec: results} reports predictive performance, computational cost, UMAP visualizations, and correlation analyses. Section~\ref{sec: discussion} discusses the implications of the results, limitations of the current study, and directions for future work.

\section{Data and study design}\label{sec: data_description}

\begin{figure}
    \centering
    \begin{subfigure}[t]{0.23\textwidth}
        \centering
        \includegraphics[height=4.5cm]{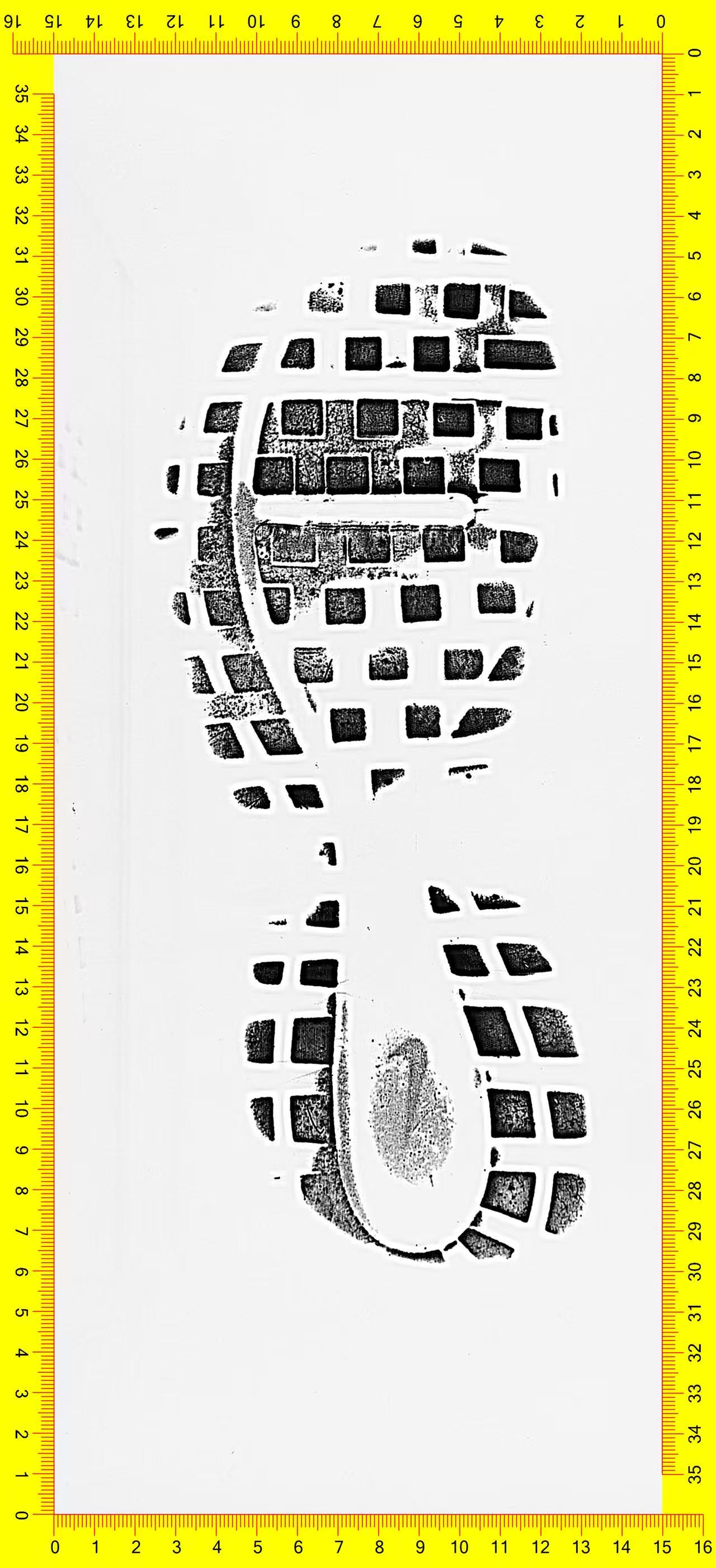}
        \caption{Male, Nike, scan 1}
        \label{fig:sample_m1}
    \end{subfigure}
    \hfill
    \begin{subfigure}[t]{0.23\textwidth}
        \centering
        \includegraphics[height=4.5cm]{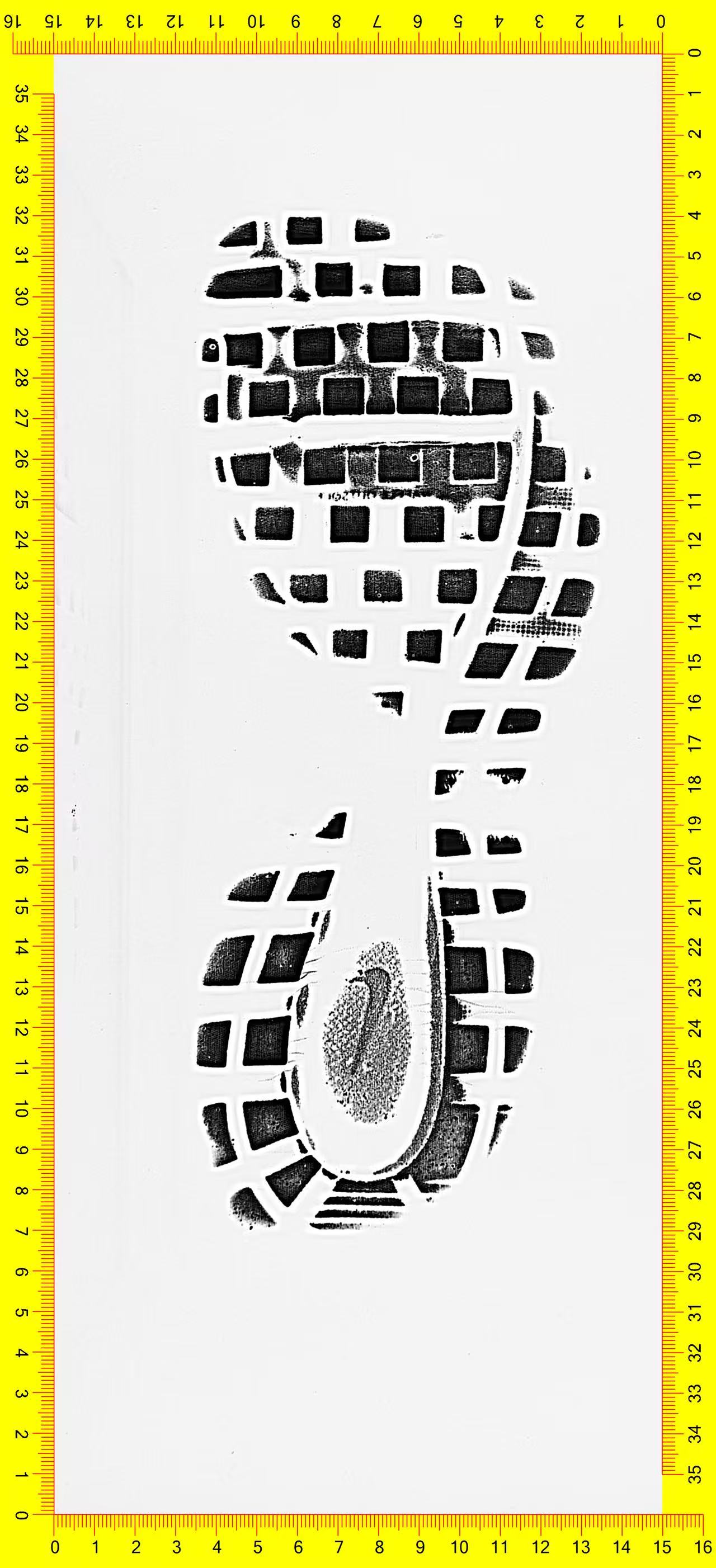}
        \caption{Same Nike, scan 4}
        \label{fig:sample_m4}
    \end{subfigure}
    \hfill
    \begin{subfigure}[t]{0.23\textwidth}
        \centering
        \includegraphics[height=4.5cm]{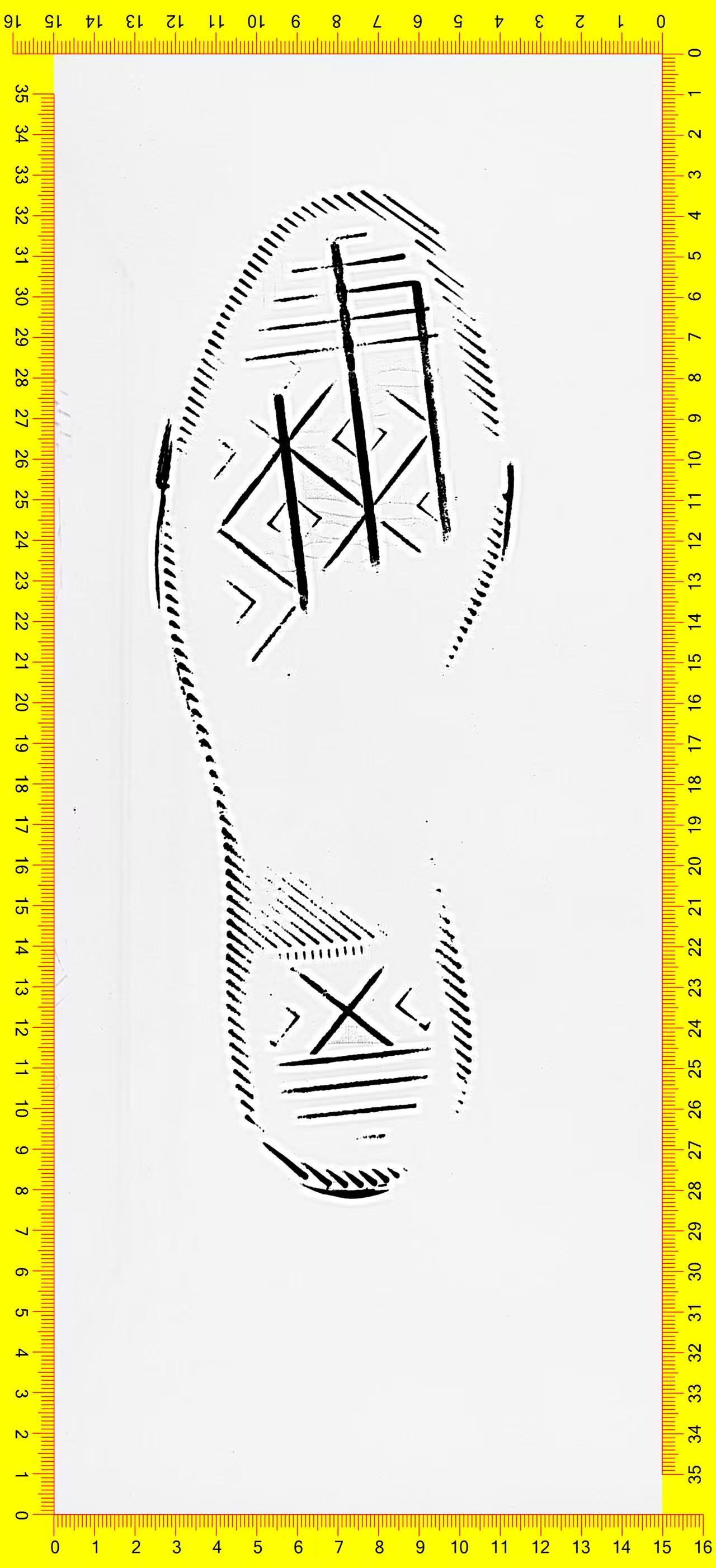}
        \caption{Female, Converse, scan 1}
        \label{fig:sample_f1}
    \end{subfigure}
    \hfill
    \begin{subfigure}[t]{0.23\textwidth}
        \centering
        \includegraphics[height=4.5cm]{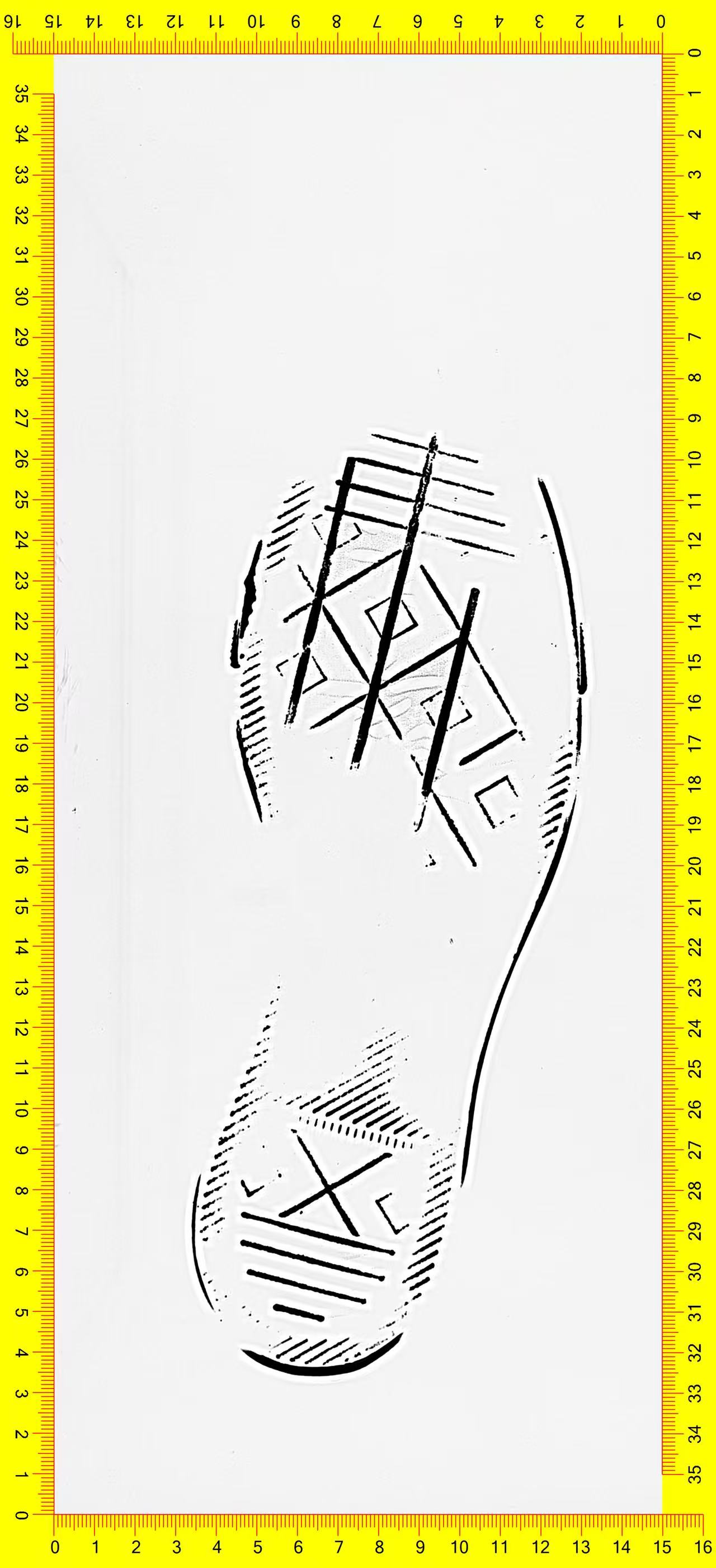}
        \caption{Same Converse, scan 4}
        \label{fig:sample_f4}
    \end{subfigure}
    \caption{Representative outsole impressions from the dataset of \citet{park2020database}. Panels (a) and (b) show two replicate scans of one male-worn Nike shoe, and panels (c) and (d) show two replicate scans of one female-worn Converse shoe. Each scan retains the yellow centimeter scale ruler from the EverOS scanner. The visible similarity between replicate scans motivates the shoe-level train/test partition used in this study.}
    \label{fig:sample_shoeprints}
\end{figure}

We analyzed the publicly available two-dimensional footwear outsole impression dataset introduced by \citet{park2020database}. This dataset contains 1500 outsole images obtained from 150 pairs of used shoes, corresponding to 300 individual shoes. Each shoe was scanned five times using an EverOS footwear scanner, yielding replicate impressions that capture within-shoe variability across repeated scans. Each image is accompanied by metadata, including the wearer's recorded sex, foot side, shoe brand, and shoe size. Since replicate scans of the same shoe are visually similar and share the same physical outsole, this dependence should be taken into account when constructing train/test partitions. Representative replicate scans are shown in Figure~\ref{fig:sample_shoeprints}.

The prediction target is the binary sex annotation recorded in the source database. Throughout the paper, we refer to the task as sex estimation from shoeprint images, with the response variable restricted to the male/female labels available in the dataset. All outsole images were resized to $224 \times 224$ pixels to match the input dimensions required by the convolutional neural network architectures considered in the study. Pixel values were rescaled according to the input requirements of each pre-trained network. These preprocessing steps produced a common image representation for all CNN-based analyses while preserving the same underlying train/test partition across model configurations.

Since each shoe was represented by five replicate scans, we partitioned the data at the shoe level rather than the image level. This prevented scans of the same physical shoe from appearing in both training and test sets, thereby reducing leakage from shoe-specific visual patterns. The partition was implemented using \texttt{GroupShuffleSplit}, which randomly forms train/test splits while keeping all observations from the same group together. Using the shoe identifier as the grouping variable, we obtained an 80\%/20\% split that was used consistently across all model configurations.

\section{Methods}\label{sec: methods}

This section describes the predictive and exploratory analyses used for sex estimation from footwear outsole impressions. We compared convolutional neural network (CNN)-based transfer learning pipelines with traditional feature-based classifiers, and then examined the relationship between CNN-derived representations and handcrafted, geometric, and metadata-derived descriptors. All models were evaluated using the shoe-level train/test partition described in Section~\ref{sec: data_description}. The study considered two broad classes of predictive approaches. The first used CNNs to learn or extract image representations directly from the outsole images. The second used a fixed set of handcrafted, geometric, and metadata-derived descriptors as inputs to traditional machine learning classifiers.

The CNN-based analyses used four ImageNet-pretrained architectures: VGG16 \citep{simonyan2014very}, ResNet-50 \citep{he2016deep}, EfficientNet-B0 \citep{tan2019efficientnet}, and MobileNet-V2 \citep{sandler2018mobilenetv2}. These architectures were selected not to exhaust the space of CNN models, but to span a range of commonly used designs with different levels of complexity and computational efficiency: VGG16 provides a conventional high-parameter baseline; ResNet-50 represents deeper residual networks; EfficientNet-B0 represents parameter-efficient compound scaling; and MobileNet-V2 represents lightweight efficient computation. This allowed us to assess whether performance patterns were specific to one network family or consistent across designs. Each architecture was initialized with ImageNet-pretrained weights, so that models started from visual filters learned on a large-scale classification task before being adapted to shoeprint images.


For each architecture, we evaluated three transfer learning strategies. The first strategy, end-to-end fine tuning, updates the full pre-trained network on the shoeprint training data and therefore tests the benefit of domain-specific adaptation. The second strategy treats the CNN as a frozen feature extractor and trains a support vector machine (SVM) on the resulting deep feature vectors, allowing us to assess whether ImageNet-derived representations are useful without updating the convolutional layers. The third strategy concatenates frozen CNN features with handcrafted and metadata-derived descriptors before SVM classification, testing whether interpretable features provide complementary information to learned image representations. Together, these strategies compare different levels of model adaptation and feature integration.

\subsection{CNN transfer learning strategies}

\textbf{\emph{Strategy 1: End-to-end fine tuning}.} In the fine tuning strategy, each ImageNet-pretrained CNN was used as an initialization for the shoeprint classification task. The original ImageNet classification layer was replaced with a binary classification head, and all network parameters were updated using the shoeprint training data. This strategy allows both the convolutional representation layers and the final classifier to adapt to visual patterns in the outsole impressions.

\textbf{\emph{Strategy 2: Frozen CNN feature extraction with SVM}.} In the frozen-feature strategy, the pretrained CNN was used as a fixed feature extractor. The convolutional base was held fixed, so its weights were not updated during training. Each shoeprint image was passed through the network, and high level activations from the final representation layer were extracted as a deep feature vector. These CNN-derived features were then used as inputs to an SVM classifier for binary sex estimation. This strategy separates representation extraction from classification. Since only the SVM classifier is trained on the shoeprint data, it is less computationally intensive than full fine tuning.

\textbf{\emph{Strategy 3: Hybrid CNN and handcrafted feature fusion}.} The hybrid strategy combined learned CNN representations with handcrafted descriptors. For each image, the frozen CNN feature vector described above was concatenated with a 29-dimensional feature vector consisting of handcrafted, geometric, and metadata-derived variables. The combined feature vector was then used as input to an SVM classifier. This strategy was included to evaluate whether manually defined descriptors provide complementary information beyond CNN-derived features. It also provides a partial connection between black-box deep representations and interpretable image or metadata variables.

\subsection{Handcrafted, geometric, and metadata-derived features}

The 29-dimensional handcrafted feature vector consisted of four groups of variables. The first group contained 20 global Haar wavelet coefficients obtained from a two level decomposition, designed to summarize multiresolution texture and structural information. The second group contained four image-level statistics: contrast, foreground-to-background ratio, contour perimeter, and frequency threshold ratio. The third group contained two metadata-derived descriptors encoding shoe brand and shoe size. The fourth group contained three geometric descriptors computed from the image: footprint length, footprint width, and pattern thickness. We distinguish these variables from CNN-derived features because they are explicitly defined before model fitting. The 29 descriptors were fixed in advance to cover complementary information types rather than being selected or optimized on the classification task, and no feature selection or dimensionality reduction was applied. This ensures that the traditional baselines reflect a reasonable, interpretable feature set rather than a task-tuned one.

\subsection{Traditional feature-based classifiers}

As traditional baselines, we trained three classifiers using only the 29-dimensional feature vector: an SVM with radial basis function kernel \citep{cortes1995support}, a random forest classifier \citep{breiman2001random}, and an XGBoost classifier \citep{chen2016xgboost}. These models were included to quantify the predictive information available from manually specified features alone and to provide benchmarks for evaluating the added value of CNN-derived image representations. The classifiers were implemented using their default hyperparameter settings from the corresponding \texttt{scikit-learn} and \texttt{xgboost} implementations.

\subsection{UMAP-based representation analysis}\label{sec: UMAP representation analysis}

To examine the relationship between learned CNN representations and interpretable image descriptors, we conducted an exploratory representation analysis using Uniform Manifold Approximation and Projection (UMAP). For each CNN architecture, feature vectors were extracted from the penultimate representation layer and projected into a five-dimensional UMAP space using $n_{\text{neighbors}} = 15$ and $\min\_ \text{dist} = 0.1$. The first two UMAP coordinates were used for visualization, while all five UMAP coordinates were used in the subsequent correlation analysis.

For the correlation analysis, we constructed a condensed nine-dimensional descriptor set. This set included the four image-level statistics used in the handcrafted feature vector (contrast, foreground-to-background ratio, contour perimeter, and frequency threshold ratio), together with brand feature, size feature, footprint length, and the first two principal components of the 20 wavelet coefficients. The wavelet principal components were obtained by standardizing the wavelet coefficients and applying principal component analysis to the wavelet block. We then computed pairwise correlations between the five UMAP coordinates and the nine interpretable descriptors for each CNN architecture. We also examined the correlation structure among the nine interpretable descriptors themselves. This analysis was intended to assess whether low-dimensional summaries of CNN representations were associated with measurable properties of the outsole impressions.

\subsection{Model training and evaluation}

All CNN models were initialized with ImageNet-pretrained weights. Fine-tuned CNNs were trained for 30 epochs using the Adam optimizer with learning rate $10^{-3}$, batch size 32, and a step learning rate scheduler. Hyperparameters were fixed in advance and were not selected using test-set performance. No separate validation set or early stopping criterion was used; the final epoch checkpoint was used for reporting.

Predictive performance was evaluated using accuracy, class-specific precision, class-specific recall (sensitivity), and F1-score \citep{hossin2015review}. To summarize computational cost, we also recorded training time and model complexity, reported through the number of model parameters.

\section{Results}\label{sec: results}

\subsection{Predictive performance}

\begin{table}[ht]
\centering
\caption{\textbf{Model performance comparison on the held-out shoe-level test partition.} Strategy 1 denotes end-to-end fine-tuning; Strategy 2 denotes frozen CNN feature extraction followed by SVM classification; Strategy 3 denotes hybrid feature fusion of frozen CNN features and the 29-dimensional feature vector followed by SVM classification. SVM denotes support vector machine; RF denotes random forest; XGBoost denotes extreme gradient boosting.}
\label{tab:model_performance}
\resizebox{\textwidth}{!}{%
\begin{tabular}{cccccccc}
\toprule[1.5pt] 
\toprule
\textbf{Model} & \textbf{Approach} & \textbf{Accuracy(\%)} & \textbf{Precision(\%)} & \textbf{Recall(\%)} & \textbf{F1\_Score(\%)} & \textbf{Parameters(m)} & \textbf{Training\_Time(s)} \\
\midrule
\multirow{3}{*}{EfficientNet-B0} & Strategy-1 & \textbf{100.00} & \textbf{100.00} & \textbf{100.00} & \textbf{100.00} & 4.01 & 1644 \\
& Strategy-2 & 93.00 & 93.13 & 93.00 & 92.98 & 4.01 & 57 \\
& Strategy-3 & 93.67 & 93.97 & 93.67 & 93.64 & 4.01 & 101 \\
\midrule
\multirow{3}{*}{MobileNetV2} & Strategy-1 & 99.67 & 99.67 & 99.67 & 99.67 & 2.23 & 1656 \\
& Strategy-2 & 90.67 & 90.71 & 90.67 & 90.67 & 2.23 & 58 \\
& Strategy-3 & 96.33 & 96.39 & 96.33 & 96.34 & 2.23 & 99 \\
\midrule
\multirow{3}{*}{ResNet50} & Strategy-1 & 99.33 & 99.34 & 99.33 & 99.33 & 23.51 & 1663 \\
& Strategy-2 & 82.00 & 85.12 & 82.00 & 81.40 & 23.51 & 60 \\
& Strategy-3 & 87.00 & 88.14 & 87.00 & 86.83 & 23.51 & 105 \\
\midrule
\multirow{3}{*}{VGG16} & Strategy-1 & 92.67 & 92.81 & 92.67 & 92.67 & 134.27 & 1682 \\
& Strategy-2 & 87.00 & 87.17 & 87.00 & 87.01 & 134.27 & 65 \\
& Strategy-3 & 75.67 & 76.86 & 75.67 & 75.17 & 134.27 & 108 \\
\midrule
- & SVM & 70.00 & 72.53 & 70.00 & 69.54 & - & 97 \\
- & RF & 69.67 & 71.49 & 69.67 & 69.36 & - & 331 \\
- & XGBoost & 65.33 & 70.74 & 65.33 & 63.72 & - & 279 \\
\bottomrule
\bottomrule[1.5pt] 
\end{tabular}}
\end{table}

Table~\ref{tab:model_performance} summarizes the predictive performance of all fifteen model configurations on the held-out shoe-level test partition. The fine-tuned CNN models (Strategy 1) achieved the strongest performance across the evaluated approaches. EfficientNet-B0 attained 100.00\% accuracy, precision, recall, and F1-score, followed by MobileNet-V2 (99.67\% accuracy), ResNet-50 (99.33\%), and VGG16 (92.67\%). Because these estimates are based on a single partition, they should be interpreted as point estimates rather than as evidence of perfect classification; variability across alternative splits is not captured here.

The frozen-feature strategy (Strategy~2) produced lower accuracy than end-to-end fine-tuning, but remained substantially stronger than the traditional feature-based baselines. EfficientNet-B0 features followed by SVM classification achieved 93.00\% accuracy, while MobileNet-V2 features achieved 90.67\%. VGG16 and ResNet-50 frozen features achieved 87.00\% and 82.00\% accuracy, respectively. These results indicate that ImageNet-pretrained CNN representations contain useful information for this task even when the convolutional layers are not updated on the shoeprint data.

The hybrid strategy (Strategy~3) combined frozen CNN features with the 29-dimensional feature vector. It improved performance over frozen CNN features alone for EfficientNet-B0, MobileNet-V2, and ResNet-50. The largest improvement was observed for MobileNet-V2, whose accuracy increased from 90.67\% under Strategy~2 to 96.33\% under Strategy~3. EfficientNet-B0 improved from 93.00\% to 93.67\%, and ResNet-50 improved from 82.00\% to 87.00\%. In contrast, the hybrid strategy reduced VGG16 performance from 87.00\% to 75.67\%, suggesting that feature fusion did not benefit all CNN architectures equally.

The traditional feature-based classifiers achieved substantially lower performance than the CNN-based pipelines. Among the three baselines, SVM performed best, with 70.00\% accuracy and 69.54\% F1-score. Random Forest achieved 69.67\% accuracy and 69.36\% F1-score, while XGBoost achieved 65.33\% accuracy and 63.72\% F1-score. These results suggest that the handcrafted, geometric, and metadata-derived descriptors contain predictive information, but less than the learned image representations extracted by CNNs.

\subsection{Computational cost and model complexity}

The results also reveal a trade-off between predictive performance and computational cost. End-to-end fine-tuning produced the highest accuracies, but required substantially longer training times, ranging from approximately 1,644 to 1,682 seconds across the four CNN architectures. The frozen-feature and hybrid strategies, in contrast, required much shorter training times, generally 57 to 108 seconds.

Model size varied considerably across architectures. VGG16 had the largest parameter count (approximately 134.27 million) but did not achieve the best predictive performance. ResNet-50 had 23.51 million parameters, EfficientNet-B0 had 4.01 million, and MobileNet-V2 had only 2.23 million. Notably, EfficientNet-B0 and MobileNet-V2 achieved the strongest fine-tuning results while using far fewer parameters than VGG16, suggesting that, within this dataset, predictive accuracy was not simply determined by model size.

The traditional feature-based classifiers had lower predictive accuracy than the CNN-based methods. Their training times were also not uniformly shorter than the frozen CNN pipelines; for example, Random Forest and XGBoost required 331 and 279 seconds, respectively, whereas frozen CNN feature extraction followed by SVM classification required approximately 57 to 65 seconds across the CNN architectures. These comparisons suggest that frozen CNN feature extraction may provide a favorable balance between accuracy and computational efficiency.

\subsection{UMAP visualization of CNN-derived representations}

\begin{figure}[H]
    \centering
    \begin{subfigure}[b]{0.48\textwidth}
        \centering
        \includegraphics[width=\textwidth]{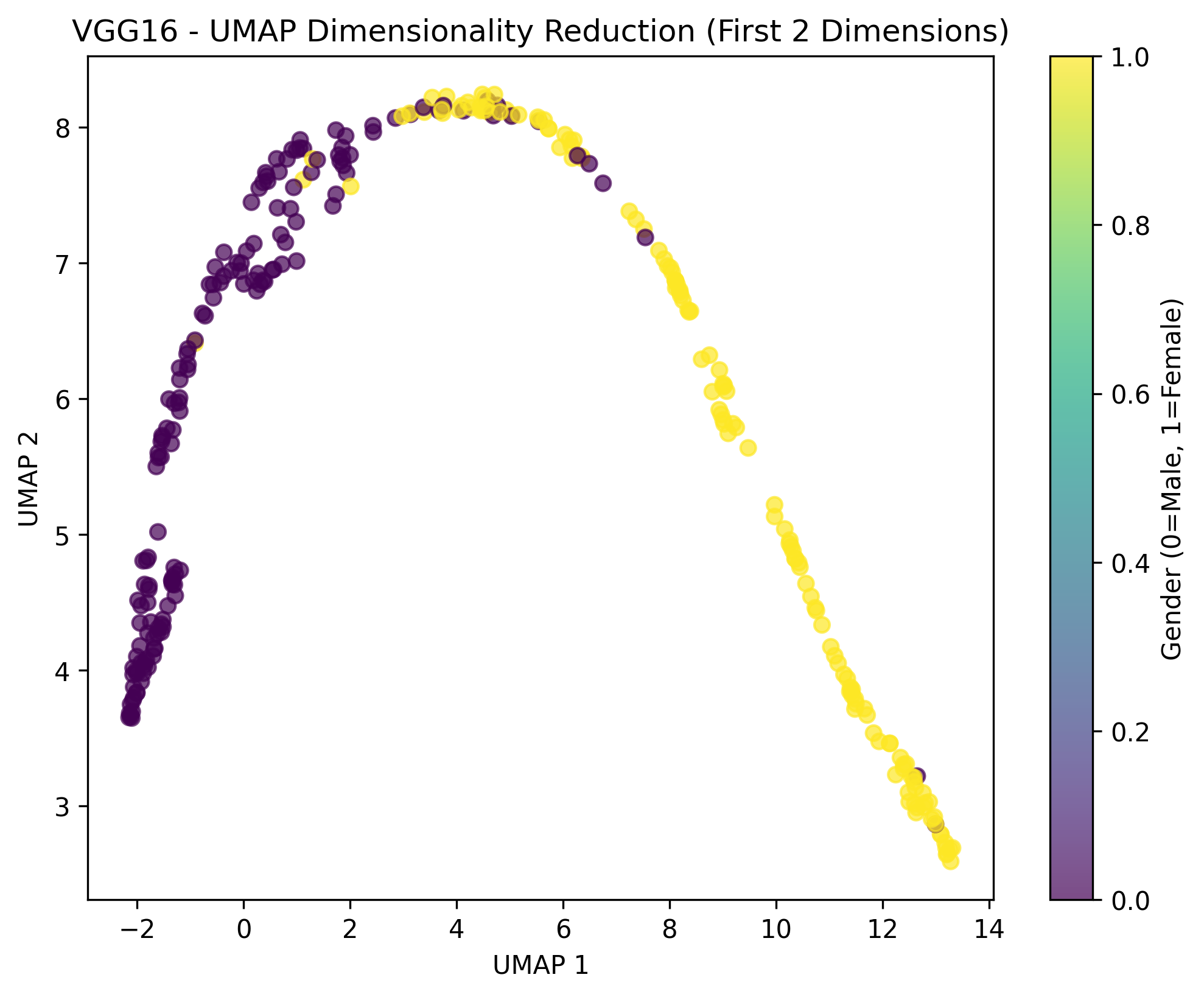}
        \caption{VGG16}
        \label{fig:umap_vgg16_en}
    \end{subfigure}
    \hfill
    \begin{subfigure}[b]{0.48\textwidth}
        \centering
        \includegraphics[width=\textwidth]{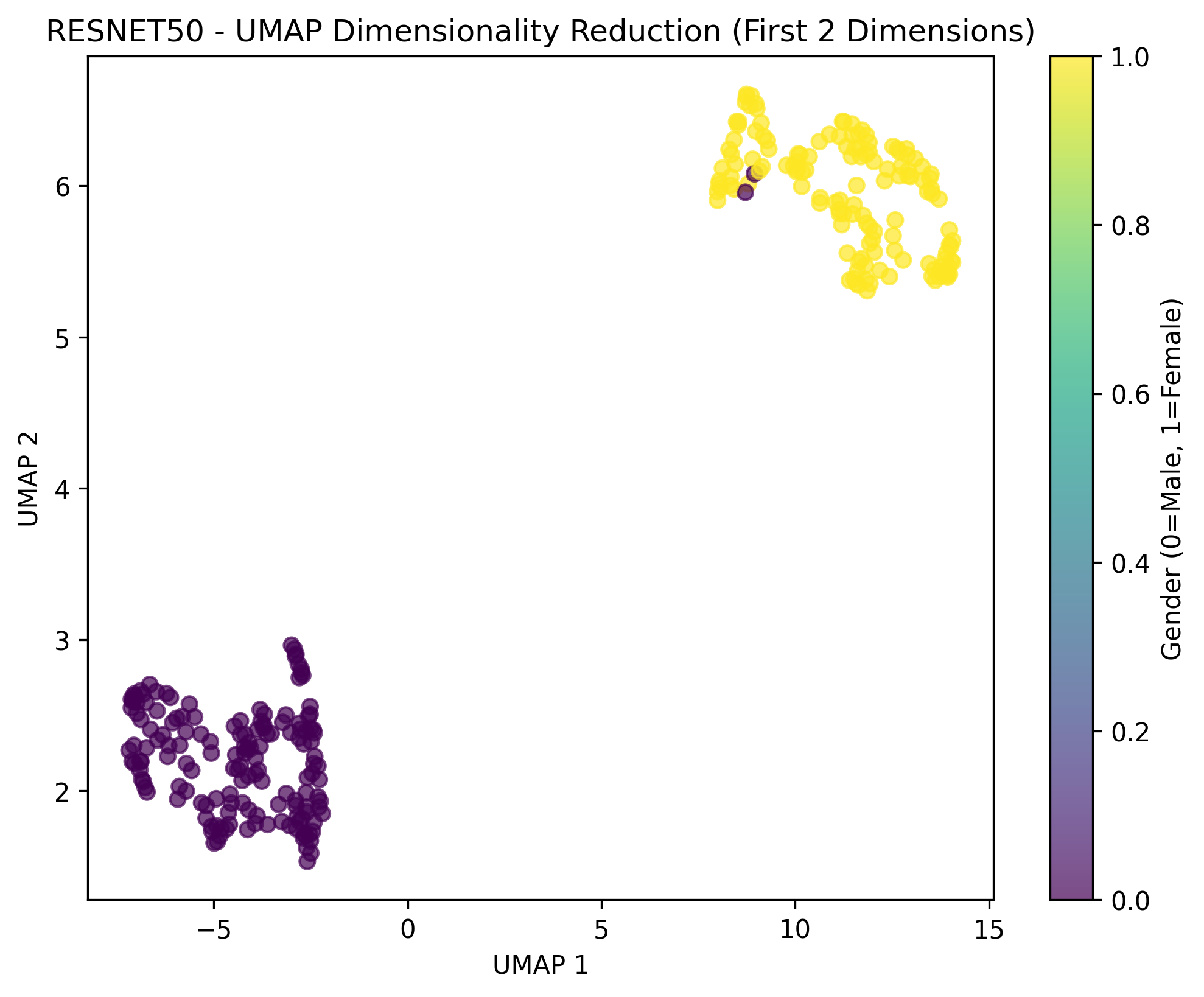}
        \caption{ResNet50}
        \label{fig:umap_resnet50_en}
    \end{subfigure}
    \vfill
    \begin{subfigure}[b]{0.48\textwidth}
        \centering
        \includegraphics[width=\textwidth]{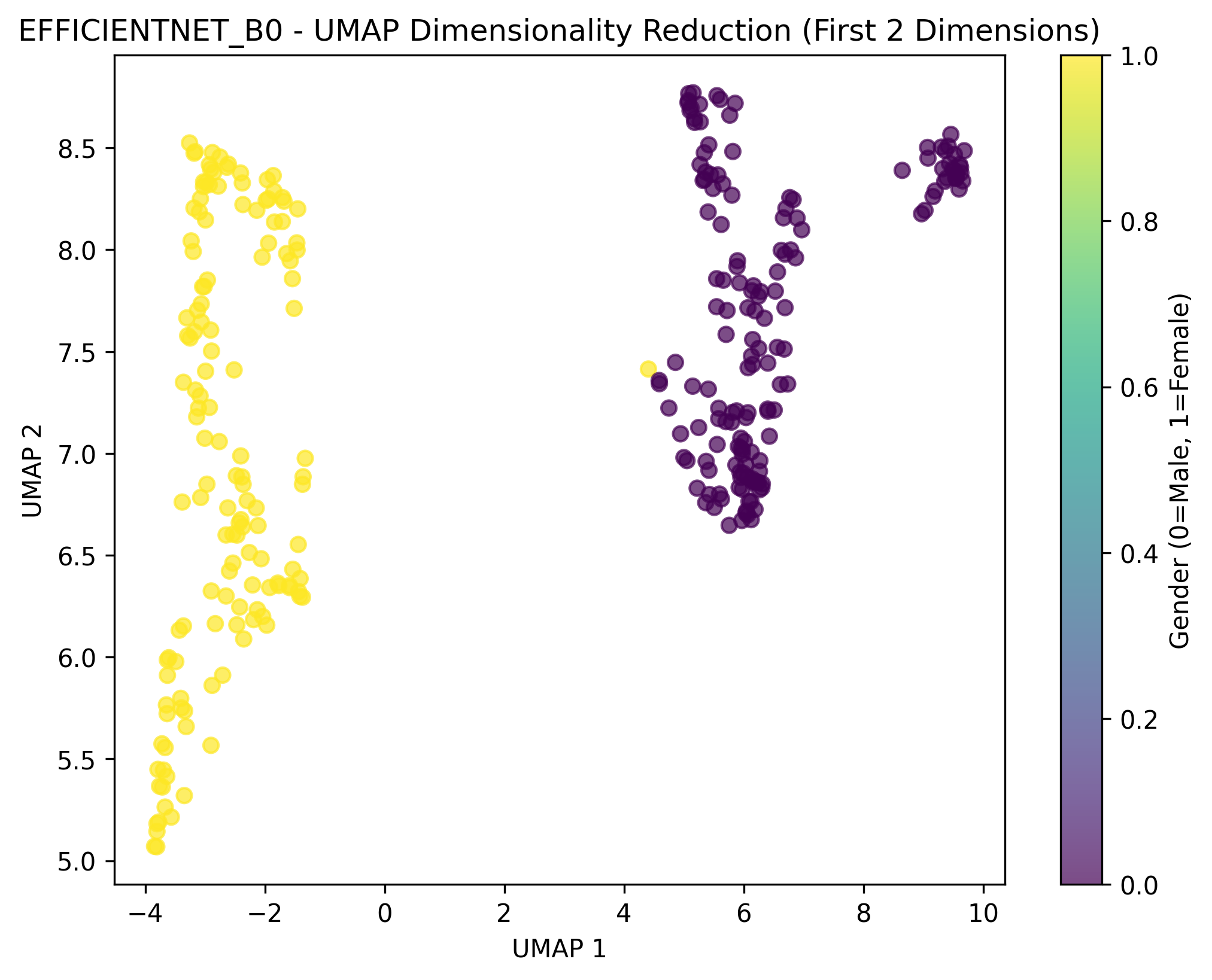}
        \caption{EfficientNet-B0}
        \label{fig:umap_efficientnet_en}
    \end{subfigure}
    \hfill
    \begin{subfigure}[b]{0.48\textwidth}
        \centering
        \includegraphics[width=\textwidth]{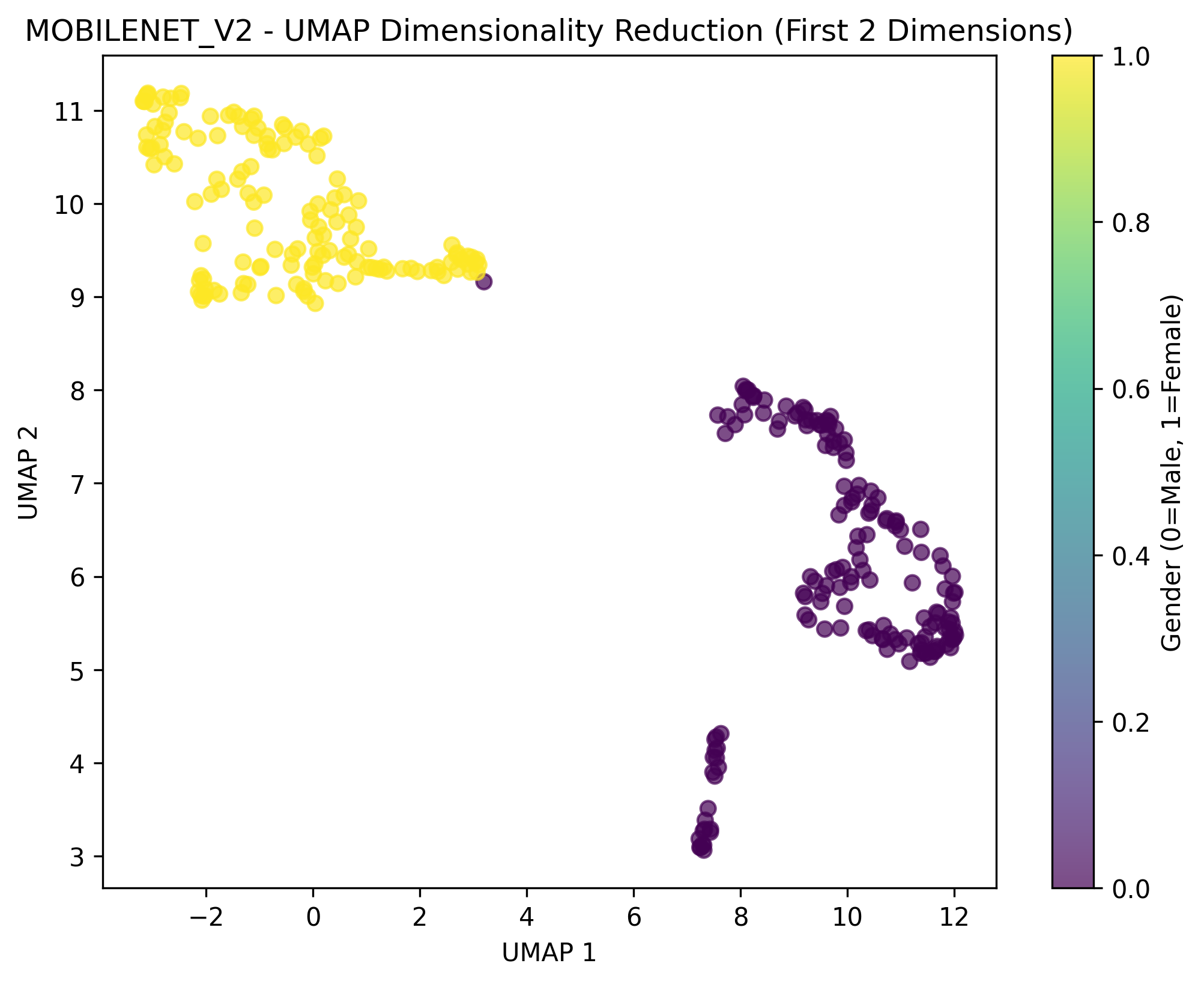}
        \caption{MobileNet-V2}
        \label{fig:umap_mobilenet_en}
    \end{subfigure}
    \caption{\textbf{UMAP visualization of CNN-derived feature representations.} The first two UMAP coordinates are shown for feature vectors extracted from VGG16, ResNet-50, EfficientNet-B0, and MobileNet-V2. Points correspond to shoeprint scans and are colored by the recorded binary sex label.}
    \label{fig:umap_results_en}
\end{figure}

Figure~\ref{fig:umap_results_en} displays two-dimensional UMAP visualizations of the CNN-derived feature representations for the four architectures. Each point corresponds to a shoeprint scan, and points are colored by the recorded binary sex label. All four architectures show visible separation between male- and female-labeled samples, though the degree of separation varies: ResNet-50 produces the most distinct clusters, while VGG16 shows the most overlap. Silhouette coefficients computed on the UMAP embedding were highest for ResNet-50 (0.72) and VGG16 (0.68); values above 0.5 indicate moderate cluster structure, though this metric is heuristic and does not directly reflect classification performance. These visualizations support the quantitative classification results by showing that CNN-derived representations encode information relevant to the sex-estimation task, and they motivate the correlation analysis in next section. However, because UMAP is a nonlinear dimensionality-reduction method, the plots should be interpreted as exploratory summaries of the feature space rather than as direct evidence of the features used by the models for classification.

\subsection{Correlation between CNN representations and auxiliary descriptors}\label{sec:correlation_cnn_descriptors}

\begin{figure}[H]
\centering
\includegraphics[width=0.8\textwidth]{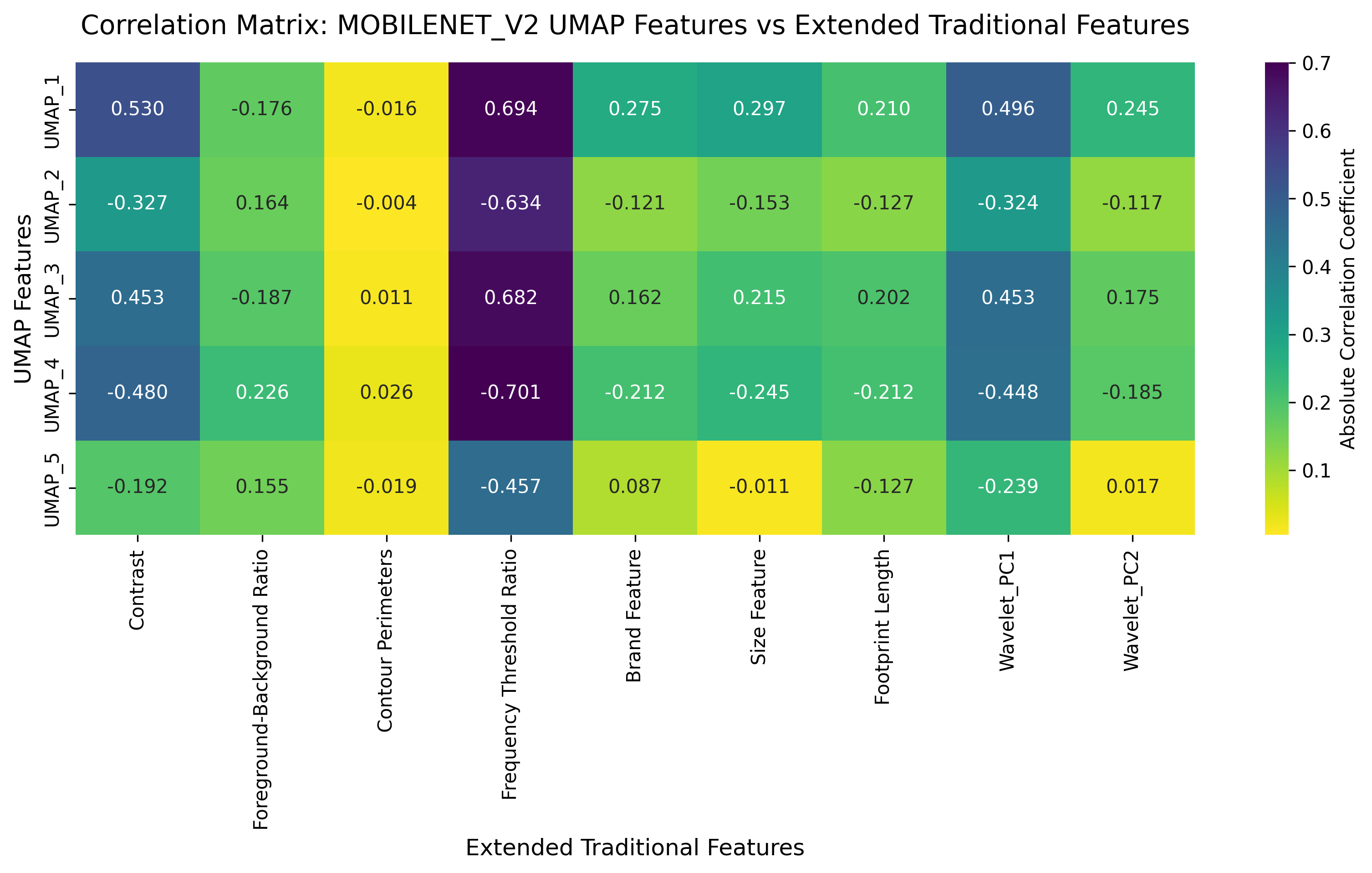}
\caption{\textbf{Correlation heatmap between MobileNet-V2 UMAP coordinates and auxiliary descriptors.} Rows correspond to the five UMAP coordinates obtained from MobileNet-V2-derived feature vectors, and columns correspond to the nine auxiliary descriptors used in the representation analysis: Contrast, Foreground-Background Ratio, Contour Perimeter, Frequency Threshold Ratio, Brand Feature, Size Feature, Footprint Length, Wavelet\_PC1, and Wavelet\_PC2. Cell values report pairwise correlation coefficients.}
\label{fig:mobilenet_correlation}
\end{figure}

\begin{figure}[H]
\centering
\includegraphics[width=0.8\textwidth]{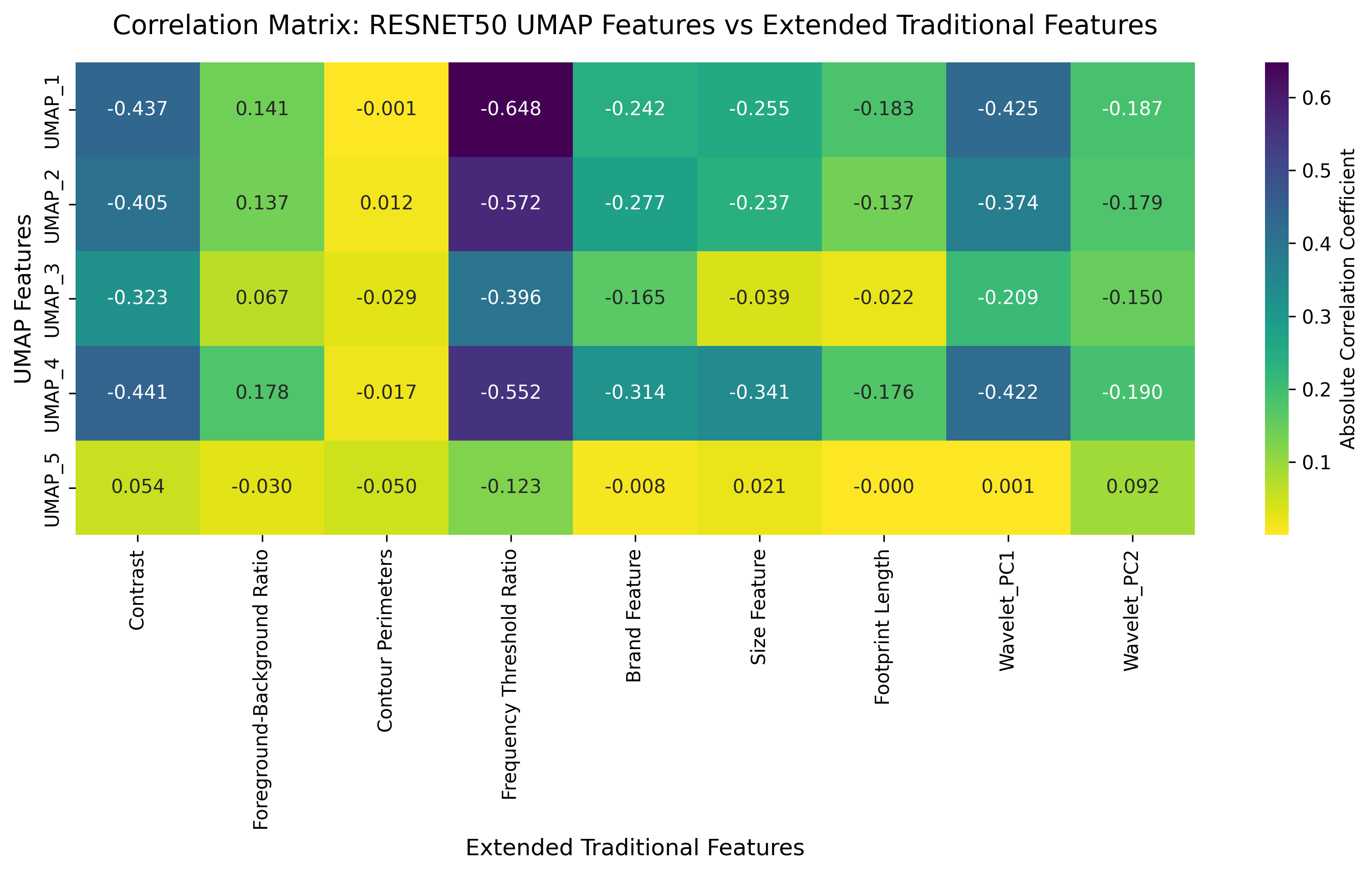}
\caption{\textbf{Correlation heatmap between ResNet-50 UMAP coordinates and auxiliary descriptors.} Rows correspond to the five UMAP coordinates obtained from ResNet-50-derived feature vectors, and columns correspond to the nine auxiliary descriptors used in the representation analysis. Cell values report pairwise correlation coefficients between the projected CNN representation and the auxiliary descriptors. The heatmap is intended as an exploratory summary of representational associations rather than a direct measure of feature importance.}
\label{fig:resnet_correlation}
\end{figure}

\begin{figure}[H]
\centering
\includegraphics[width=0.8\textwidth]{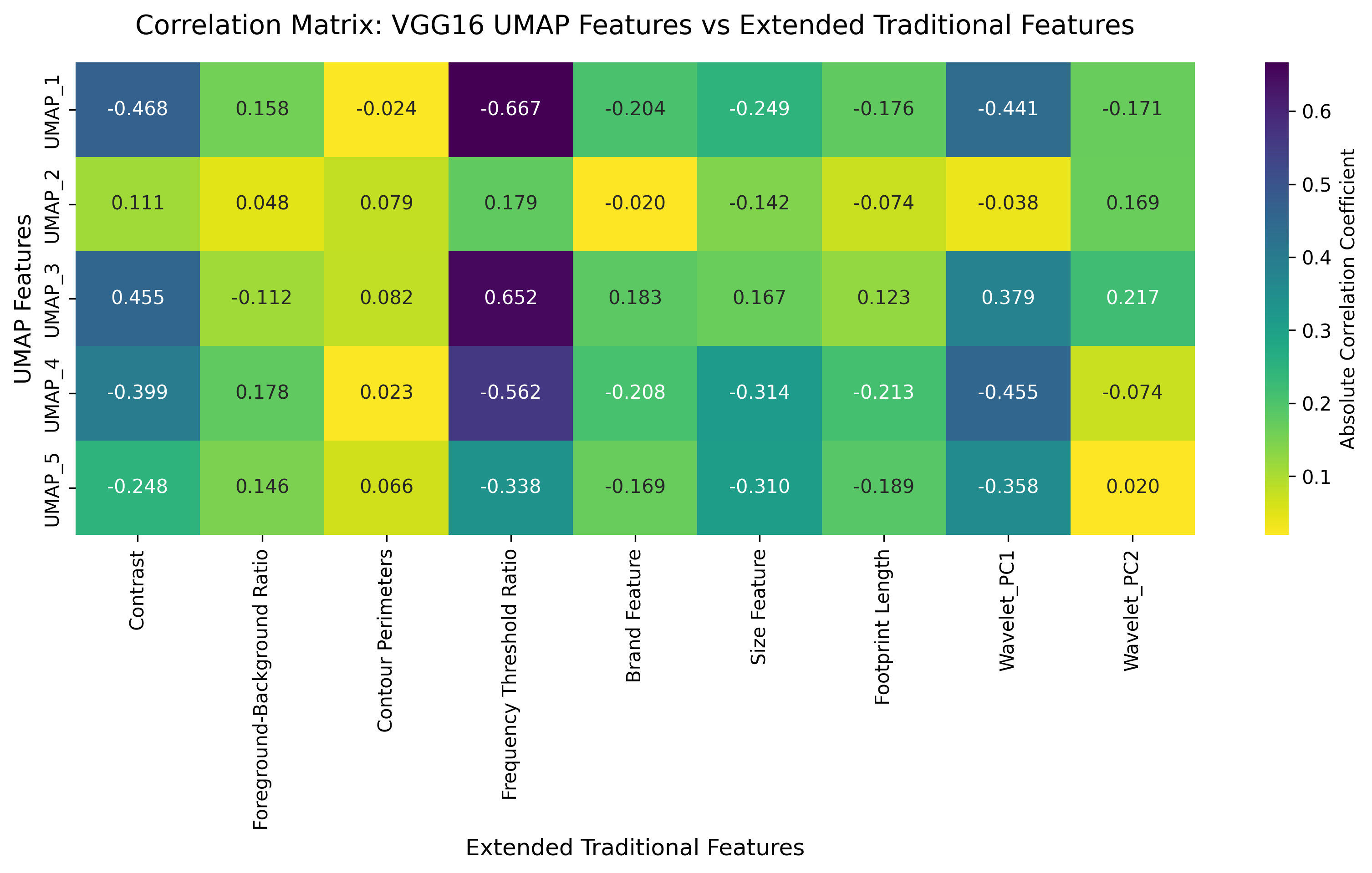}
\caption{\textbf{Correlation heatmap between VGG16 UMAP coordinates and auxiliary descriptors.} Rows correspond to the five UMAP coordinates obtained from VGG16-derived feature vectors, and columns correspond to the nine auxiliary descriptors used in the representation analysis. Cell values report pairwise correlation coefficients. The heatmap shows how low-dimensional summaries of the VGG16 representation are associated with quantifiable image-level, geometric, and metadata-derived descriptors.}
\label{fig:vgg_correlation}
\end{figure}

\begin{figure}[H]
\centering
\includegraphics[width=0.8\textwidth]{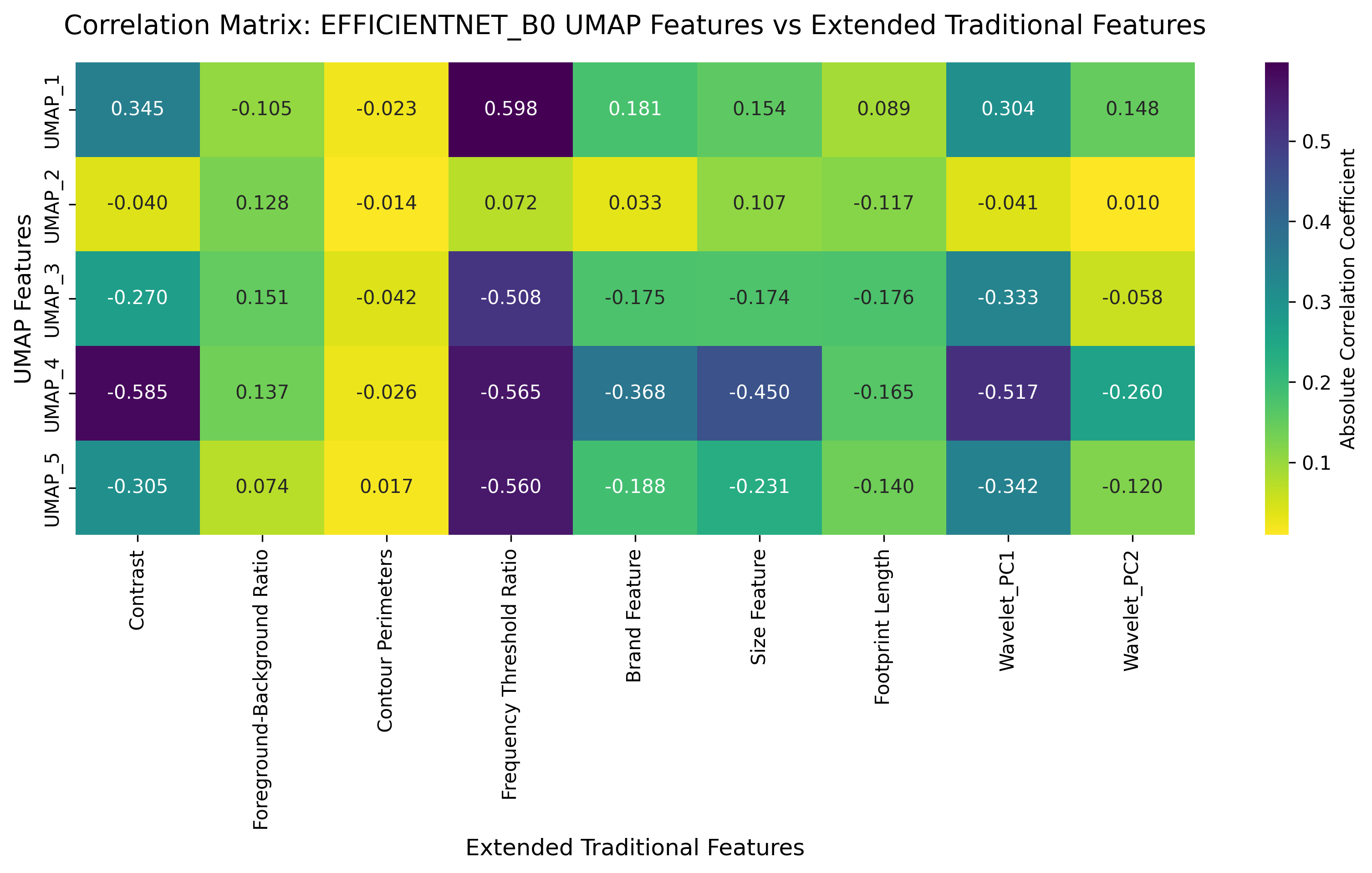}
\caption{\textbf{Correlation heatmap between EfficientNet-B0 UMAP coordinates and auxiliary descriptors.} Rows correspond to the five UMAP coordinates obtained from EfficientNet-B0-derived feature vectors, and columns correspond to the nine auxiliary descriptors used in the representation analysis. Cell values report pairwise correlation coefficients. The heatmap provides an exploratory comparison between the projected EfficientNet-B0 representation and the handcrafted, geometric, and metadata-derived descriptors.}
\label{fig:efficientnet_correlation}
\end{figure}

To further examine the relationship between CNN-derived representations and quantifiable descriptors, we correlated the five-dimensional UMAP projections with the nine-dimensional descriptor set described in Section~\ref{sec: UMAP representation analysis}. Figures~\ref{fig:mobilenet_correlation}–\ref{fig:efficientnet_correlation} show the resulting correlation heatmaps for MobileNet-V2, ResNet-50, VGG16, and EfficientNet-B0, respectively. Across the four CNN architectures, Frequency Threshold Ratio showed the most consistent association with the UMAP coordinates, with absolute correlations generally ranging from 0.60 to 0.70.  In the MobileNet-V2 heatmap, for example, Frequency Threshold Ratio had a positive correlation of 0.694 with UMAP\_1 and a negative correlation of -0.701 with UMAP\_4. Contrast was also associated with the UMAP representation, with a correlation of 0.530 with UMAP\_1 in the MobileNet-V2 analysis. Wavelet\_PC1 appeared as another secondary contributor across the heatmaps. The strength of association varied across architectures: MobileNet-V2 had the largest mean absolute correlation between its UMAP-projected features and the nine descriptors (0.255), followed by VGG16 (0.223), EfficientNet-B0 (0.210), and ResNet-50 (0.206). These values suggest that the learned representations differ in how closely their low-dimensional summaries align with the handcrafted, geometric, and metadata-derived descriptors.

The correlation results should be interpreted cautiously. They indicate statistical association between UMAP-projected CNN features and measurable descriptors, but they do not establish that the CNNs causally rely on any specific descriptor when making predictions. Nevertheless, the repeated association with Frequency Threshold Ratio, Contrast, and Wavelet\_PC1 suggests that frequency- and contrast-related properties of outsole images are reflected in the learned representations.

\subsection{Correlation structure among auxiliary descriptors}

\begin{figure}[H]
\centering
\includegraphics[width=0.8\textwidth]{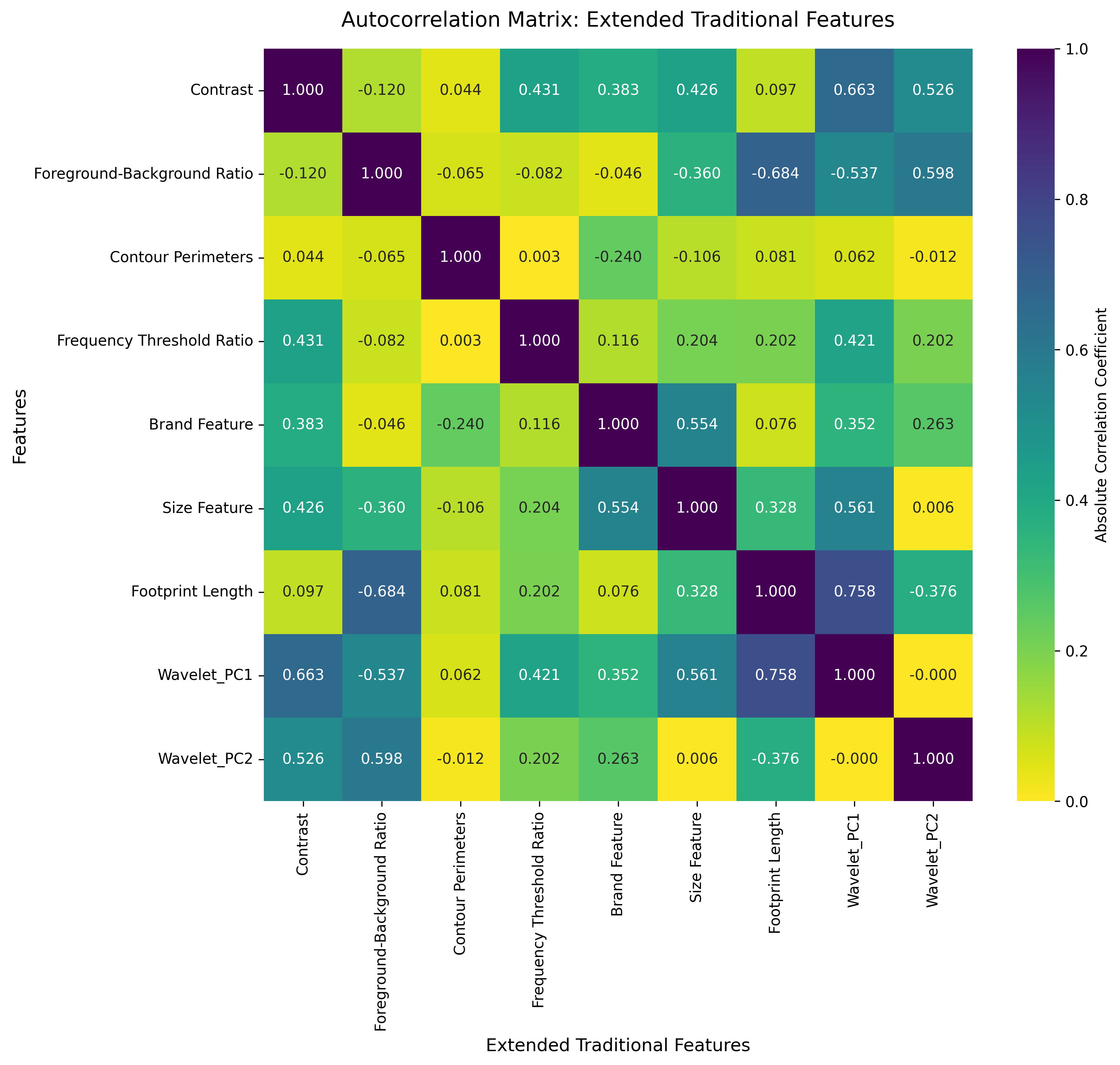}
\caption{\textbf{Correlation matrix among the nine auxiliary descriptors.}
The heatmap summarizes pairwise correlations among the image-level statistics, metadata-derived variables, footprint length, and wavelet principal components used in the representation analysis.}
\label{fig:traditional_autocorrelation}
\end{figure}

Figure~\ref{fig:traditional_autocorrelation} shows the correlation matrix among the nine auxiliary descriptors. This analysis assessed the degree of dependence among the interpretable variables themselves. The heatmap indicates moderate correlation among several descriptors: approximately 22.2\% of off-diagonal feature pairs had absolute correlations greater than 0.5. Some of the strongest associations involved wavelet and geometric information. Wavelet\_PC1 was positively correlated with Contrast ($r=$0.663) and with Footprint Length ($r=$0.758). The metadata-derived variables were also correlated with image-derived descriptors: Brand Feature was correlated with Contrast ($r=$0.383) and Wavelet\_PC1 ($r=$0.352), while Size Feature was correlated with Wavelet\_PC1 ($r=$0.561). These correlations indicate that the auxiliary descriptors are not mutually independent. Therefore, associations between CNN-derived UMAP coordinates and a particular descriptor may also reflect relationships with other correlated descriptors. This dependence should be considered when interpreting the correlation heatmaps in Section~\ref{sec:correlation_cnn_descriptors}.

\section{Discussion}\label{sec: discussion}

Our results show that CNN-based transfer learning substantially outperforms traditional feature-based classifiers for sex estimation from footwear outsole impressions. Under the shoe-level partition, fine-tuned CNNs achieved the highest test accuracies, with EfficientNet-B0, MobileNet-V2, and ResNet-50 all performing strongly, whereas the traditional baselines trained only on the 29-dimensional feature vector achieved substantially lower accuracy. These findings suggest that the raw outsole images contain discriminative information not fully captured by the manually specified descriptors. End-to-end fine-tuning outperformed frozen feature extraction, consistent with the role of fine-tuning in adapting ImageNet-pretrained representations to a new image domain; frozen-feature and hybrid strategies nonetheless remained substantially stronger than the traditional baselines and required far less training time, suggesting that pretrained CNNs can serve as effective feature extractors when computational resources are limited. The hybrid strategy improved performance for three of the four architectures, indicating that handcrafted, geometric, and metadata-derived descriptors may provide complementary information, although the decrease observed for VGG16 shows that feature fusion does not uniformly help. Across architectures, predictive performance was not simply determined by model size: EfficientNet-B0 and MobileNet-V2 achieved the highest fine-tuning accuracies while using far fewer parameters than VGG16.

The representation analysis provides additional insights into the learned feature spaces. UMAP visualizations showed visible separation between male- and female-labeled samples across all four architectures, and the correlation analysis linked low-dimensional summaries of CNN features to quantifiable descriptors, most consistently Frequency Threshold Ratio, followed by Contrast and Wavelet\_PC1. These associations suggest that frequency- and contrast-related image properties are reflected in the learned representations, providing a bridge between high-dimensional CNN features and measurable image characteristics. However, because the correlations are based on UMAP-projected representations rather than the original feature spaces or input perturbations, they indicate association rather than causal feature use, and the auxiliary descriptors were themselves moderately correlated, so any single association may partly reflect relationships with other descriptors. The analysis should therefore be interpreted as exploratory representation analysis rather than definitive model explanation.

From a forensic perspective, these results support the potential value of CNN-based methods as investigative screening tools, but not as definitive evidence about an individual: their most plausible use is as an auxiliary triage tool, where an estimated sex label could help prioritize leads during early-stage investigation. Operational use would require additional validation, uncertainty quantification, and evaluation on more realistic forensic images. Several limitations qualify these findings. First, all images come from a single public dataset collected under controlled conditions using one scanner type and acquisition protocol, so the reported accuracies should be interpreted as held-out performance within this dataset rather than as evidence of operational validity. Second, performance was evaluated using a single 80\%/20\% partition, which does not quantify variability across alternative splits; future work should use shoe-level cross-validation or repeated group-wise splits. Third, the response variable is restricted to the binary male/female annotation in the source database, and the results should not be interpreted as making claims about sex or gender categories beyond those available. Finally, external validation on independently collected datasets such as mock crime-scene or casework-like impressions involving partial prints, variable substrates, and degradation, would be necessary before drawing conclusions about practical deployment, and should preserve the same leakage-aware design principles used here. 


\section{Acknowledgement}\label{acknowledgments}
The authors
thank Soyoung Park and Alicia Carriquiry for making the footwear
outsole impression dataset publicly available, without which this
study would not have been possible.

\section{Disclosure statement}\label{disclosure-statement}

The authors report there are no competing interests to declare.

\section{Data availability statement}\label{data-availability-statement}

Data are from the public footwear outsole impression dataset reported by \citet{park2020database}.

\bibliographystyle{chicago}
\bibliography{references}

\end{document}